\documentclass[10pt,twocolumn]{article}

\usepackage[letterpaper,margin=1in,columnsep=0.25in]{geometry}

\usepackage{amsmath,amssymb}
\usepackage{booktabs}
\usepackage{algorithmic}
\usepackage{algorithm}
\usepackage[hidelinks]{hyperref}
\usepackage{url}
\usepackage[expansion=false]{microtype}
\usepackage{graphicx}
\usepackage{natbib}
\usepackage{xcolor}
\usepackage{enumitem}
\usepackage{multirow}
\usepackage{tabularx}
\usepackage{tikz}
\usepackage{pgfplots}
\pgfplotsset{compat=1.17}
\usetikzlibrary{arrows.meta,positioning,shapes.geometric,fit,calc}

\definecolor{xfpblue}{HTML}{2563EB}
\definecolor{xfpred}{HTML}{DC2626}
\definecolor{xfpgreen}{HTML}{16A34A}
\definecolor{xfpgray}{HTML}{6B7280}
\definecolor{xfporange}{HTML}{EA580C}
\definecolor{abstractbg}{HTML}{F8FAFC}
\definecolor{abstractborder}{HTML}{CBD5E1}

\title{\textbf{XFP: Quality-Targeted Adaptive Codebook Quantization}\\[3pt]
\textbf{with Sparse Outlier Separation for LLM Inference}}

\author{
\textbf{Thomas Witt}\\[4pt]
Gemini Stiftung Leipzig\\[2pt]
{\small \texttt{witt@gemini-stiftung.de}}\\[6pt]
{\small Code \& Data:
\url{https://github.com/flash7777/vllm/tree/multiquant}}
}

\date{}

\begin{document}

\twocolumn[%
\maketitle

\vspace{-0.5em}

\noindent\fcolorbox{abstractborder}{abstractbg}{%
\begin{minipage}{\dimexpr\textwidth-2\fboxsep-2\fboxrule}
\vspace{0.5em}
\centerline{\large\textbf{Abstract}}
\vspace{0.4em}
We introduce XFP, a dynamic weight quantizer for LLM inference that
inverts the conventional quantization workflow: the operator specifies
reconstruction quality floors on per-channel cosine similarity (one
strict floor for attention and shared experts, one lazy floor for
routed-expert MoE); XFP determines codebook size, outlier budget, and
packing per layer automatically---no Hessian, no calibration data, no
manual bit-width selection.

Each weight matrix is decomposed into a sparse fp16 outlier residual
and a dense sub-byte index tensor into a per-group learned codebook
(group size 128, configurable). The same auto-select frontend, the
same outlier path, and the same fused decode kernel serve two
codebook storage modes: \textbf{V2} (per-channel Lloyd codebook) and
\textbf{V2a} (shared library of $L\!=\!32$ codebooks per layer,
indexed per group). V2 is the headline configuration; V2a is an
alternative trading codebook precision for $\sim$1~KB constant
per-CTA SMEM, useful where the activation-cache budget is tight.

Three contributions: \textbf{(C1)}~The XFP quantizer with
quality-targeted two-threshold auto-select, converging to $\sim$3.97
effective bits on Qwen3.5-122B-A10B and to $\sim$3.4 on
Qwen3.5-397B-A17B under the H-Process (\S\ref{sec:exp-397b}).
\textbf{(C2)}~Two codebook storage modes sharing one auto-select
frontend and one fused decode kernel, covering 30B--397B on
workstation-class GPUs without NVFP4 Tensor Cores.
\textbf{(C3)}~The H-Process: a quality-driven iteration over the two
cosine thresholds that fits a 397B MoE into $2\times$96~GB at
single-stream throughput exceeding INT4 with routed-expert pruning,
without removing experts.

On Qwen3.5-122B-A10B under V2, XFP achieves 138~tok/s single-stream
decode on workstation hardware (RTX PRO 6000 Blackwell, TP=2) at
94.49\% GSM8K strict-match (3 seeds, $n\!=\!3{,}957$), and is 49\%
faster than Marlin INT4 at TP=1. On Qwen3.5-397B-A17B under the
H-Process at H1.5 ($\tau_\text{strict}\!=\!0.96$,
$\tau_\text{lazy}\!=\!0.93$, $\sim$3.4 effective bits), the same engine
delivers 100.9~tok/s long-output decode at 66.72\% GSM8K strict-match
on the full 1{,}319-problem set (single seed at submission;
multi-seed evaluation in progress).
\vspace{0.4em}
\end{minipage}%
}

\vspace{1.2em}
]

\section{Introduction}

It was still snowing in Leipzig when we first tried to serve
Qwen3.5-397B-A17B---a 397-billion-parameter Mixture-of-Experts
model---on two freshly delivered RTX PRO 6000 Blackwell GPUs with 96~GB
each. The uncompressed weights exceed 700~GB. At INT4 the model fits, but
barely: KV cache is starved, context length is capped, and the quality
loss from uniform 4-bit quantization is substantial. We tried aggressive
quantization, expert pruning, and combinations of both. The result was
consistent: below a certain compression threshold, output quality
collapses. The question that drove this work was not \emph{how} to
quantize---GPTQ, AWQ, and others have solved that---but \emph{why the
quality loss at 4 bits is as large as it is}.

Today's quantization landscape offers two paradigms: \textbf{linear}
quantization (the INT-$N$ family), which maps weights to a uniform grid
scaled by a single factor, and \textbf{logarithmic} quantization (NVFP4
and related formats), which uses a fixed non-uniform codebook whose
entries follow a floating-point-like spacing. Both add per-group or
per-block scaling to absorb magnitude variation. Both apply the same
grid---linear or logarithmic---to every layer, every expert, every
projection matrix in the model. Weights, activations, KV cache:
everything is forced through the same representational template.

Once this uniformity is acknowledged, the quality loss is unsurprising.
A weight distribution with catastrophic outliers at $49\sigma$ (as we
observed in GLM-4.7-Flash attention) and a weight distribution with a
tight Gaussian bulk (as in the same model's routed experts) receive the
same codebook. The outlier-dominated layer wastes most of its
representational budget on a range that contains 0.3\% of its weights;
the well-behaved layer is over-provisioned with entries it does not need.

We then looked at NVIDIA's NVFP4---announced with considerable fanfare
as a ``4-bit floating-point'' format with dedicated Tensor Core support
on Blackwell. Upon inspection, the mechanism is straightforward: a fixed
16-entry codebook lookup. The same 16 values for every layer, every
model, every architecture. The ``floating point'' in the name refers to
the codebook entries, not to the representation. On our RTX PRO 6000
(SM120), the dedicated Tensor Core path for NVFP4 is unavailable---that
hardware acceleration is reserved for datacenter Blackwell (SM100+). So
we were left with a format that promises hardware support we cannot use
and a universal codebook that ignores the distributions we can see.

This led to a straightforward idea: if the underlying primitive is a
codebook lookup, why not learn the codebook \emph{individually}---one per
layer, one per group of output channels---from the actual weight
distribution? And if the codebook is learned, why fix its size to 16
entries when some layers need only 4 or 8? XFP is our answer: a
quantizer that learns a per-layer codebook of adaptive size, separates
outliers into a sparse residual, and lets the operator specify quality
rather than bit width.

Three structural problems in existing approaches frame the design:

\paragraph{P1---The Outlier Problem.}
Transformer weight matrices exhibit heavy-tailed distributions. A small
fraction of weights (typically 0.3--2\%) occupy a magnitude range far
exceeding the bulk. Linear INT-$N$ quantization accommodates this range in
a single scale factor, degrading resolution for the dense central region.
At $N\!=\!2$, outlier-driven scaling leaves effectively one usable non-zero
level for the bulk.

\paragraph{P2---The MoE Heterogeneity Problem.}
In Mixture-of-Experts (MoE) models, individual experts and component
classes exhibit distinct weight distributions. Routed experts are
empirically more tolerant of coarse quantization than attention or shared
experts; a single global quality floor cannot exploit this asymmetry.

\paragraph{P3---The Codebook-Storage Problem.}
A per-channel learned codebook occupies kernel shared memory in
proportion to the output dimension; on workstation-class GPUs with
99~KB SMEM/CTA this caps the activation-row cache and forces an early
split-$K$, limiting the reachable model size at single-stream decode.

\subsection{Design Principle: Quality as Constraint}

Existing quantization formats require the operator to choose a target
bit width and accept whatever quality results. XFP inverts this: the
operator specifies two \textbf{reconstruction quality floors} on
per-channel cosine similarity---a strict floor for attention and
shared-expert projections, a lazy floor for routed-expert MoE---and
XFP determines the minimum bit width per layer that meets the relevant
floor. Weights not captured by the codebook are separated into a sparse
fp16 outlier residual.

Three consequences follow: (1)~bit width becomes a per-layer derived
quantity; (2)~the outlier residual is not a heuristic add-on but the
structural complement guaranteeing the quality floor; (3)~the same
algorithm adapts to different models without manual tuning, with the
threshold gap as the single operator-tunable knob trading memory
against quality on a fixed deployment target.

\subsection{MultiQuant}

XFP is developed within MultiQuant, a component-level quantization engine
for vLLM~\citep{kwon2023vllm}. MultiQuant maintains independent precision
policies per component class (attention, routed experts, shared experts,
LM heads) through registry-based dispatch. This per-component architecture
enables XFP's sub-4-bit operation: at 3 or 2 bits, the margin between
``works'' and ``collapses'' is determined by the worst-case component
class.

\subsection{Contributions}

\textbf{(C1) A Dynamic, Quality-Targeted Quantizer.} The operator
specifies two quality floors; XFP determines everything else---codebook
size, outlier budget, packing---per layer, with a single auto-select
frontend driving both supported codebook storage modes
(\S\ref{sec:format}--\ref{sec:kernel}).

\textbf{(C2) Two Codebook Storage Modes, One Engine.} A per-channel
Lloyd codebook (\textbf{V2}) and a per-group shared-library codebook
(\textbf{V2a}) share the same outlier path, the same two-threshold
auto-select, and the same fused decode kernel skeleton---covering the
30B--397B parameter range on workstation-class GPUs without NVFP4
Tensor Cores.

\textbf{(C3) The H-Process: Constrained Compression on 397B.}
A quality-driven iteration over the two cosine thresholds fits
Qwen3.5-397B-A17B (512 routed experts/layer) on a
$2\times$96~GB workstation at $\sim$3.4 effective bits, retaining
the full expert population, at 100.9~tok/s single-stream decode and
66.72\% GSM8K strict-match on the full 1{,}319-problem set
(\S\ref{sec:exp-397b}).

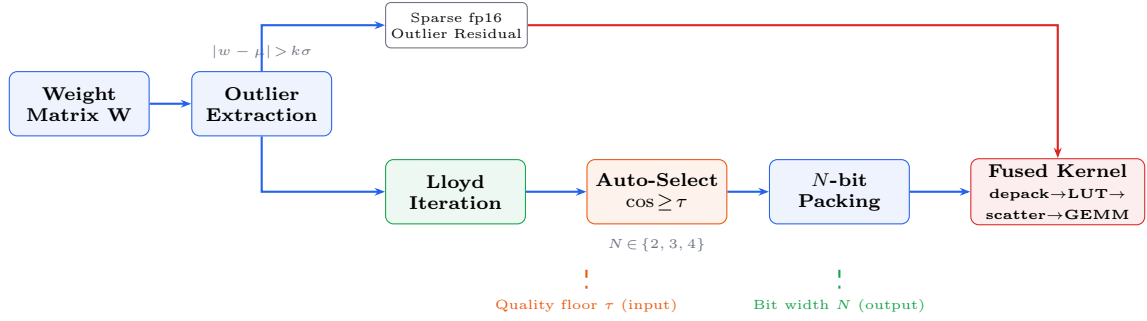
\begin{figure*}[t]
\centering
\begin{tikzpicture}[
    node distance=0.4cm and 0.55cm,
    block/.style={rectangle, draw=xfpblue, fill=xfpblue!8, rounded corners=3pt,
        minimum height=0.85cm, minimum width=1.85cm, align=center,
        font=\scriptsize\bfseries, inner sep=2pt},
    smallblock/.style={rectangle, draw=xfpgray, fill=white, rounded corners=2pt,
        minimum height=0.6cm, minimum width=1.4cm, align=center, font=\tiny, inner sep=2pt},
    arrow/.style={-{Stealth[length=4pt]}, thick, xfpblue},
    darrow/.style={-{Stealth[length=4pt]}, thick, xfpred},
    label/.style={font=\tiny\itshape, xfpgray},
]
\node[block] (input) {Weight\\Matrix $\mathbf{W}$};

\node[block, right=of input] (split) {Outlier\\Extraction};
\node[label, above=0.05cm of split] {$|w-\mu|\!>\!k\sigma$};

\node[smallblock, above right=0.3cm and 0.7cm of split] (outlier) {Sparse fp16\\Outlier Residual};
\node[block, below right=0.3cm and 0.7cm of split, fill=xfpgreen!8, draw=xfpgreen] (lloyd) {Lloyd\\Iteration};

\node[block, right=0.8cm of lloyd, fill=xfporange!8, draw=xfporange] (auto) {Auto-Select\\$\cos\!\geq\!\tau$};
\node[label, below=0.05cm of auto] {$N\!\in\!\{2,3,4\}$};

\node[block, right=of auto] (pack) {$N$-bit\\Packing};

\node[block, right=0.8cm of pack, fill=xfpred!8, draw=xfpred,
    minimum width=2.3cm] (kernel) {Fused Kernel\\{\tiny depack$\to$LUT$\to$}\\{\tiny scatter$\to$GEMM}};

\draw[arrow] (input) -- (split);
\draw[arrow] (split) |- (outlier);
\draw[arrow] (split) |- (lloyd);
\draw[arrow] (lloyd) -- (auto);
\draw[arrow] (auto) -- (pack);
\draw[arrow] (pack) -- (kernel);
\draw[darrow] (outlier) -| (kernel);

\draw[dashed, xfporange, thick] ([yshift=-0.6cm]auto.south west) -- ++(0,-0.25cm)
    node[below, font=\tiny, xfporange] {Quality floor $\tau$ (input)};
\draw[dashed, xfpgreen, thick] ([yshift=-0.6cm]pack.south) -- ++(0,-0.25cm)
    node[below, font=\tiny, xfpgreen] {Bit width $N$ (output)};
\end{tikzpicture}
\caption{\textbf{XFP pipeline overview.} The operator specifies a quality
floor $\tau$; XFP determines everything else. Outlier extraction separates
high-magnitude weights into a sparse fp16 residual. Lloyd iteration learns
a per-layer codebook on the cleaned bulk distribution. Auto-select
(\textcolor{xfporange}{Algorithm~1}) tests candidate codebook sizes and
picks the minimum $N$ meeting $\tau$. The fused decode kernel reconstructs
weights at inference via codebook gather + outlier scatter-add.}
\label{fig:pipeline}
\end{figure*}

\section{Related Work}
\label{sec:related}

\paragraph{Linear quantization.}
Round-to-nearest (RTN) with per-group or per-channel scaling is the
simplest post-training quantization scheme. GPTQ~\citep{frantar2022gptq}
improves on RTN via column-wise quantization with Hessian-based error
compensation, yielding substantially better perplexity at the cost of
calibration data and $O(d^2)$ compute per layer.
AWQ~\citep{lin2023awq} identifies salient weight channels via activation
statistics and applies per-channel scaling before quantization; it avoids
the Hessian but still requires calibration data. Both operate in the
linear INT-$N$ paradigm---the codebook is implicitly a uniform grid, and
outliers are handled through scale adjustment or error compensation, not
separation.

Intel's AutoRound~\citep{cheng2024autoround} takes a different approach:
it optimizes weight rounding decisions via signed gradient descent on a
small calibration set, achieving near-lossless INT4 quality in minutes.
Independent benchmarks~\citep{marie2025nvfp4} show AutoRound INT4
consistently matching or exceeding NVIDIA's own NVFP4 format in accuracy
on identical models---a notable result given that NVFP4 has dedicated
hardware support.

\paragraph{Codebook-based and non-linear quantization.}
AQLM~\citep{egiazarian2024aqlm} represents each weight vector as a sum
of entries from multiple learned codebooks (typically 2), achieving strong
results at 2 bits but requiring expensive calibration-data-driven
multi-codebook optimization.
QuIP/QuIP\#~\citep{chee2023quip,tseng2024quip2} use random orthogonal
transforms (incoherence processing) to spread outlier energy across all
weights before quantization, achieving excellent 2-bit results at the cost
of Hessian computation and a decode-time transform.
QTIP~\citep{tseng2024qtip} extends QuIP\# with trellis-coded
quantization for near-optimal rate-distortion at 2 bits.

SqueezeLLM~\citep{kim2023squeezellm} and SpQR~\citep{dettmers2023spqr}
are conceptually closest to XFP: both decompose weight matrices into a
sparse high-precision component (for outliers or salient weights) and a
dense low-precision component. The key difference is the identification
criterion: SpQR uses the Hessian to determine which weights are ``sensitive'';
XFP uses the weight distribution itself ($|w - \mu| > k\sigma$),
eliminating the Hessian entirely. SpQR's dense component uses a fixed
linear quantizer; XFP replaces it with a per-group learned codebook
(group size 128 by default, configurable).

\paragraph{Expert pruning for MoE deployment.}
For very large MoE models that exceed a target memory envelope, an
orthogonal family of methods reduces the routed-expert population
rather than the per-weight precision. REAP~\citep{lasby2024reap}
identifies low-utility experts from a calibration set and removes them
permanently. The approach is effective but introduces a sensitivity to
the calibration distribution: experts deleted offline may have been
load-bearing for a workload the calibration set did not represent. The
authors' REAP-it-Yourself (RIY) workflow addresses this by letting
operators run the identification pass on their own traffic, producing
deployment-specific expert profiles; a paper on RIY is in preparation
separately. We compare to REAP/RIY as an alternative deployment path
for the 397B case (\S\ref{sec:exp-397b}): XFP+H trades per-weight bits
for memory; REAP/RIY trades expert population for memory. Both are
valid responses to the same constraint.

\paragraph{Hardware-specific formats and NVIDIA's quantization ecosystem.}
NVIDIA has invested heavily in format-level quantization support across
several generations. FP8 (E4M3/E5M2) is supported on Hopper (SM90+) and
Blackwell; NVFP4~\citep{nvidia2024nvfp4} adds a fixed 16-entry codebook
with dedicated Tensor Core acceleration on datacenter Blackwell (SM100+).
MXFP4/MXFP6~\citep{ocp2023mx} (OCP Microscaling) provide block-scaled
floating-point with shared exponents.

On workstation Blackwell (SM120/121), NVFP4 support was initially absent
despite NVIDIA's launch-day marketing. The CUTLASS NVFP4 MoE kernels
failed on SM120 due to TMA work-sharing grouped-GEMM initialization
errors; FlashInfer did not compile for the \texttt{sm120f} family target
required for the FP4 conversion PTX instruction; and vLLM's backend
selection did not recognize SM120 as NVFP4-capable, falling back silently
to Marlin INT4. Community-contributed patches to vLLM, FlashInfer, and
the CUTLASS dispatch logic were required before NVFP4 checkpoints could
run at all on RTX PRO 6000 or DGX Spark hardware. Even after these
fixes, the native NVFP4 Tensor Core path on SM120 underperforms the
Marlin W4A16 fallback on MoE workloads, as the grouped-GEMM
specialization remains incomplete.

The software side is equally active: NVIDIA's TensorRT Model
Optimizer~\citep{nvidia2024modelopt} bundles SmoothQuant, AWQ, and---since
2025---NVFP4 PTQ and QAT pipelines. Despite this investment, independent
evaluations consistently find that NVFP4's fixed universal codebook
underperforms calibration-aware methods (AutoRound, AWQ) in accuracy at
the same bit width. The format's advantage is throughput via Tensor
Core acceleration---an advantage unavailable on the workstation tier
where memory pressure is most acute.

\paragraph{Inference kernels and practical deployment.}
Marlin~\citep{frantar2024marlin} is the most optimized INT4 dequantization
kernel for NVIDIA GPUs, using a 128-thread warp-level tile structure with
cooperative shared-memory staging. It serves as our primary throughput
reference. The llama.cpp project~\citep{gerganov2023llamacpp} provides
block-level codebook quantization (Q2\_K through Q6\_K) for CPU inference
---the closest practical precedent to per-block codebook quantization at
deployment scale, though with fixed rather than learned codebooks and
without outlier separation.

XFP is integrated into vLLM~\citep{kwon2023vllm} via the MultiQuant
engine, which provides per-component-class quantization dispatch
(\S\ref{sec:related}, MultiQuant).

\paragraph{Positioning.}
XFP trades the rate-distortion optimality of Hessian-aware methods for a
different optimization target: where GPTQ/AQLM/QuIP\# produce the best
reconstruction at a \emph{given} bit width, XFP takes a quality floor as
input and finds the \emph{minimum} compression meeting it---automatically,
per layer, without calibration. Where NVFP4 offers hardware-accelerated
throughput with a universal codebook, XFP offers software-defined
throughput with a learned codebook---and achieves higher throughput on
workstation hardware where NVFP4's Tensor Core path is unavailable.

\begin{table}[t]
\centering
\caption{Positioning of XFP vs.\ existing methods.}
\label{tab:positioning}
\scriptsize
\setlength{\tabcolsep}{3pt}
\begin{tabular}{@{}lcccccc@{}}
\toprule
 & GPTQ & AQLM & QuIP\# & SpQR & NVFP4 & \textbf{XFP} \\
\midrule
Learned CB    & --         & \checkmark & --         & --         & --         & \checkmark \\
Hessian-free  & --         & --         & --         & --         & \checkmark & \checkmark \\
Calib.-free   & --         & --         & --         & --         & \checkmark & \checkmark \\
Outlier sep.  & --         & --         & --         & \checkmark & --         & \checkmark \\
Auto bit sel. & --         & --         & --         & --         & --         & \checkmark \\
HW-agnostic   & \checkmark & \checkmark & \checkmark & \checkmark & --         & \checkmark \\
MoE-native    & --         & --         & --         & --         & --         & \checkmark \\
\bottomrule
\end{tabular}
\end{table}

\section{The XFP Format}
\label{sec:format}

\subsection{Decomposition}

XFP represents a weight matrix $\mathbf{W}$ as:
\begin{equation}
\mathbf{W} \approx \mathbf{W}_\text{outlier} + \text{Codebook}[\mathbf{Q}_\text{bulk}]
\label{eq:decomp}
\end{equation}
where $\mathbf{W}_\text{outlier}$ is a sparse fp16 matrix of high-magnitude
weights at their original positions, and $\mathbf{Q}_\text{bulk}$ is a
dense $N$-bit index matrix into a learned Codebook. The XFP engine
supports two codebook storage modes behind a single auto-select frontend
and a single fused decode kernel skeleton:
\begin{itemize}[leftmargin=1.1em, itemsep=1pt, topsep=2pt]
\item \textbf{V2: per-channel codebook.} One independent fp16 codebook
  of $2^N$ entries per output channel, learned via per-channel Lloyd
  iteration. Supports $N \in \{2,3,4,5,6\}$.
\item \textbf{V2a: shared codebook library.} A small fp16 library of $L$
  codebooks (default $L\!=\!32$) per layer; each weight group (default
  $128$ contiguous weights along the output dimension) is assigned the
  best-fitting library entry. Supports $N \in \{2,4\}$
  (\S\ref{sec:format-v2a}).
\end{itemize}
V2 is the production format for the 30B--122B reference experiments
(\S\ref{sec:experiments}); V2a is the format that brought a 397B MoE
into single-node 2$\times$96~GB deployment (\S\ref{sec:exp-397b}).

\subsection{Storage Layout}

Per layer, V2 stores: packed bulk indices ($N$ bits/weight), codebook
(fp16, $2^N$ entries per output channel), and outlier triples (row index
int64, column index int64, value fp16 = 18 bytes/outlier). At the
default 2\% outlier cap, outlier storage adds 0.36 bytes/weight.

V2a replaces the per-channel codebook by a per-layer library of $L$
codebooks (fp16, $2^N$ entries each, $L\!=\!32$) plus a 5-bit
group$\to$library assignment, one assignment per 128-weight group along
the output dimension. The per-group overhead is amortized across 128
weights: $\sim$$0.31$ bits/weight for the assignment, group-scale, and
mid-point. Across $N\!\in\!\{2,4\}$ this gives effective bit widths of
$\sim$$2.31$ (V2a at $N\!=\!2$) and $\sim$$4.31$ (V2a at $N\!=\!4$) per
parameter, before the sparse outlier residual.

\subsection{Sub-Byte Packing}

\begin{table}[t]
\centering
\caption{XFP sub-byte packing schemes. V2 supports all listed $N$; V2a
imposes the additional constraint that
$\text{LANES\_PER\_GROUP} = \text{group\_size}\,/\,\text{values\_per\_word}$
be an integer with $\text{warpSize} \bmod \text{LANES\_PER\_GROUP} = 0$.
At the default group size of 128 this selects $N\!\in\!\{2,4\}$:
$N\!=\!3$ fails $128 \bmod 10 \neq 0$; $N\!=\!5,6$ exceed the warp-wide
LUT budget (\S\ref{sec:format-v2a}).}
\label{tab:packing}
\small
\setlength{\tabcolsep}{4pt}
\begin{tabular}{@{}cccccccc@{}}
\toprule
$N$ & Vals/word & Word & Used & Reserve & V2 & V2a \\
\midrule
2 & 16 & uint32 & 32 & 0 & \checkmark & \checkmark \\
3 & 10 & uint32 & 30 & 2 & \checkmark & --- \\
4 & 8  & uint32 & 32 & 0 & \checkmark & \checkmark \\
5 & 3  & uint16 & 15 & 1 & \checkmark & --- \\
6 & 5  & uint32 & 30 & 2 & \checkmark & --- \\
\bottomrule
\end{tabular}
\end{table}

\subsection{V2a: Shared Codebook Library}
\label{sec:format-v2a}

V2a is an alternative storage mode in which the per-channel codebook is
replaced by a shared library at the layer level. Where V2 stores one
codebook of $2^N$ entries per output channel, V2a stores a library of
$L$ codebooks (default $L\!=\!32$) per layer and a 5-bit group$\to$library
assignment per 128-weight group. The motivation is SMEM efficiency on
workstation-class GPUs (99~KB opt-in SMEM per CTA on SM120/121, less than
half the budget of data-center SMs): the V2 per-channel codebook occupies
SMEM in proportion to the output dimension, while a V2a library of
$L\!=\!32$, $N\!=\!4$ codebooks takes $32 \cdot 16 \cdot 2$~B $=$ 1~KB
regardless of dimension. The freed budget extends the activation-row
cache from $K\!\leq\!8{,}192$ (V2 without split-$K$) to
$K\!\leq\!32{,}768$ on SM120/121.

V2 with split-$K$ dispatch handles $K\!>\!8{,}192$ as well at a 10--20\%
single-stream latency cost (\S\ref{sec:kernel}); V2a is the alternative
that avoids the split-$K$ overhead at the cost of slightly lower
codebook precision per group. On the 397B Front~B deployment
(\S\ref{sec:exp-397b}), V2 with split-$K$ is the headline configuration;
V2a serves the same workload with equivalent functionality and lower
single-stream throughput, and we report V2 numbers for clarity.

\paragraph{Library learning.}
Per layer, we collect the per-channel Lloyd codebooks that V2 would
produce, then run a second Lloyd pass over the codebooks themselves to
condense them into a library of $L\!=\!32$ representative codebooks
(\emph{LibFit}). Each 128-weight group is then re-indexed to the closest
library entry under a per-group scale and mid-point.

\paragraph{Lane geometry.}
The decode kernel walks 128-weight groups with
$\text{LANES\_PER\_GROUP} = 128 / \text{values\_per\_word}$ threads of
the warp per group, and $\text{CB\_PER\_ITER} = 32 /
\text{LANES\_PER\_GROUP}$ groups per warp iteration. For $N\!=\!4$ this
gives $\text{LANES\_PER\_GROUP}\!=\!16$, $\text{CB\_PER\_ITER}\!=\!2$;
for $N\!=\!2$, $\text{LANES\_PER\_GROUP}\!=\!8$,
$\text{CB\_PER\_ITER}\!=\!4$. $N\!=\!3$ fails the integer constraint at
$g\!=\!128$ ($128 \bmod 10 \neq 0$) and would require a separate
$g\!=\!80$ pack path (deferred); the group size itself is a tunable
parameter not further explored in this work.

\section{Encoding and Auto-Selection}
\label{sec:encoding}

\subsection{Outlier Extraction}

For each weight matrix $\mathbf{W}$: compute $\mu, \sigma$; extract
weights where $|w - \mu| > k\sigma$ (default $k\!=\!4.0$, capped at 2\%)
into a sparse fp16 residual; replace extracted positions with $\mu$ in the
bulk.

\subsection{Learned Codebook via Lloyd Iteration}

Initialize $2^N$ codebook entries via CDF-uniform quantiles of the bulk
distribution; iterate 20 rounds of assignment (nearest-neighbor) and
centroid update (conditional mean). The codebook is stored as fp16 per
output channel.

\subsection{Quality-Targeted Auto-Selection}

\begin{algorithm}[t]
\caption{XFP Auto-Select (two-threshold)}
\label{alg:autoselect}
\begin{algorithmic}[1]
\REQUIRE Weight matrix $\mathbf{W}$, layer class $c$, candidate set
  $\mathcal{N}$, strict floor $\tau_\text{strict}$, lazy floor
  $\tau_\text{lazy} \leq \tau_\text{strict}$
\STATE Extract outliers, compute $\mathbf{W}_\text{bulk}$
\STATE $\tau \gets \tau_\text{lazy}$ if $c \in \text{MoE}_\text{routed}$
  else $\tau_\text{strict}$
\FOR{$N \in \mathcal{N}$ in ascending order}
  \STATE $\text{cb} \gets \text{LearnCodebook}(\mathbf{W}_\text{bulk}, 2^N)$
    \COMMENT{V2: per-channel Lloyd; V2a: LibFit}
  \STATE $\hat{\mathbf{W}} \gets \text{Reconstruct}(\text{cb},
    \mathbf{W}_\text{outlier})$
  \STATE $\text{cos}_i \gets \text{cos\_sim}(\mathbf{W}_{i,:},
    \hat{\mathbf{W}}_{i,:}) \;\forall i$
  \IF{$\text{median}(\text{cos}) \geq \tau$}
    \RETURN $N$
  \ENDIF
\ENDFOR
\RETURN $\max(\mathcal{N})$ \COMMENT{fallback}
\end{algorithmic}
\end{algorithm}

Auto-mode tests codebook sizes in ascending order and returns the first
achieving the active quality floor. The candidate set is
$\mathcal{N}\!=\!\{2,3,4\}$ for V2 and $\mathcal{N}\!=\!\{2,4\}$ for V2a
(\S\ref{sec:format-v2a}). The gate uses \textbf{per-channel cosine
similarity}, aggregated via \textbf{median}---not mean (sensitive to
outlier channels) or min (overly conservative at $N\!=\!2$).

\paragraph{Two thresholds, two regimes.}
A single global threshold (the earlier XFP convention, $\tau\!=\!0.98$)
sufficed for 35B and 122B under V2 because $\mathcal{N}$ contained the
intermediate option $N\!=\!3$: routed experts landed at xfp3, attention
escalated to xfp4, no class was forced to choose between coarse and
excessive precision. V2a removes the $N\!=\!3$ slot (packing constraint,
\S\ref{sec:format-v2a}); on the 397B MoE this collapses auto-select to
a binary choice between $\sim$2.31 and $\sim$4.31 effective bits per
parameter for every layer class. Routed experts are empirically more
robust to coarse quantization than attention or shared experts, but
with a single threshold the binary choice forces 100\% xfp4 (memory
infeasible at 2$\times$96~GB) or near-100\% xfp2 (math collapses).

The two-threshold split decouples the regimes. $\tau_\text{strict}$
governs attention, linear-attention and shared-expert projections;
$\tau_\text{lazy}$ governs the routed-expert path. The threshold gap
controls the routed-expert xfp2/xfp4 mix and is the single parameter
operators tune to trade memory against quality on a fixed deployment
target (\S\ref{sec:exp-397b}, Table~\ref{tab:cos-sweep}). The strict
floor defaults to $0.96$ and the lazy floor to $0.93$; we configure
them via the environment variables \texttt{XFP\_MIN\_COS\_STRICT} and
\texttt{XFP\_MIN\_COS\_LAZY}. The two-threshold mechanism is engine-wide:
V2 and V2a use the same dispatch.

\paragraph{Why cosine similarity, not MSE.}
MSE is magnitude-sensitive; cosine similarity is direction-sensitive. In
a GEMM ($\mathbf{y} = \mathbf{x} \cdot \mathbf{W}_\text{col}$),
directional preservation matters more---activations absorb magnitude
scaling via learned norms. Empirically, XFP2 on routed experts has
$12\times$ higher MSE than XFP4 with identical math accuracy.

\paragraph{MoE sampling.}
For MoE layers with $E$ experts, auto-select runs on 4 concatenated
experts (not the full population). Offline validation on three models
confirms: 0\% disagreement on GLM (64 experts) and Qwen-35B (256),
11.5\% on Qwen-122B (256)---with zero math impact from disagreements
due to the sparse outlier path absorbing the error. We retain the
4-expert sample for V2a; on Qwen-397B (512 experts/layer) the
auto-select cost is dominated not by sampling but by Lloyd convergence:
the V2a MoE-LibFit benefits from a higher iteration budget
($\texttt{XFP\_MOE\_LLOYD\_ITERS}\!=\!20$ vs the V2 default of 5), since
under-converged centroids systematically under-estimate xfp2 quality
and bias auto-select toward xfp4.

\subsection{Auto-Selection Results}

\begin{table}[t]
\centering
\caption{Auto-mode convergence on GLM-4.7-Flash.}
\label{tab:glm-auto}
\small
\begin{tabular}{@{}lcl@{}}
\toprule
Component & Layers & Auto-selected \\
\midrule
Attention     & 235 & 233$\times$xfp3, 2$\times$xfp4 \\
Routed (64$\times$) & 46 groups & 46$\times$xfp3 \\
Shared        & 92  & 92$\times$xfp3 \\
\midrule
\textbf{Total} & \textbf{$\sim$375} & \textbf{$\sim$3.0 eff.\ bits} \\
\bottomrule
\end{tabular}
\end{table}

On Qwen3.5-122B, auto-mode selects XFP3 for GatedDeltaNet and shared
layers but XFP4 for routed experts ($\sim$3.97 eff.\ bits). The algorithm
is identical; different models produce different---and
appropriate---profiles.

\subsection{Encoding Cost}

Auto-mode bit-width selection (Algorithm~\ref{alg:autoselect} on 4 sampled
experts per MoE layer) completes in under one minute. Full quantize-on-load
--- Lloyd codebook fitting (V2) or LibFit (V2a), outlier extraction, and
sub-byte packing for all layers --- takes $\sim$10 minutes for
Qwen3.5-35B-A3B and $\sim$25 minutes for Qwen3.5-122B-A10B under V2, and
$\sim$30 minutes for Qwen3.5-122B-A10B and $\sim$55 minutes for
Qwen3.5-397B-A17B under V2a, on an RTX PRO 6000. Packed weights are
cached to disk; subsequent loads are I/O-bound. By comparison, GPTQ
requires comparable or longer calibration times with additional
data-dependency.

\section{Fused Decode Kernel}
\label{sec:kernel}

The decode kernel fuses four operations per layer forward pass:
(1)~depack $N$-bit indices to int8; (2)~codebook gather from shared
memory; (3)~sparse outlier scatter-add (fp16$\to$bf16); (4)~bf16 GEMM
via Tensor Cores. The codebook is loaded into SMEM once per thread
block, making the kernel bit-width-portable: the same gather path serves
$N\!=\!2$ through $N\!=\!6$ in V2 and $N\!\in\!\{2,4\}$ in V2a.

The kernel operates natively in bf16 for both activation input and output,
eliminating the bf16$\to$fp16$\to$bf16 conversion overhead that earlier
prototypes showed added 124\% per call. It is wrapped as a
\texttt{torch.library.custom\_op} for \texttt{torch.compile} compatibility
and CUDA Graph capture.

\paragraph{Activation-row caching.}
For $M\!=\!1$ decode (the dominant autoregressive regime), the kernel
caches the full $K$-dimensional activation row in shared memory once per
CTA, then iterates output groups with codebook lookups against the cached
row. This eliminates repeated global-memory reads of the
activation---the dominant traffic at $M\!=\!1$. The reachable $K$ is
bounded by the shared-memory budget per CTA after the codebook itself is
loaded, which on SM120/121 is 99~KB of opt-in SMEM. Under V2 the
per-channel codebook scales with the output dimension; once an
activation row of $K\!=\!8{,}192$ in bf16 is resident, the
remaining SMEM is fully consumed. For $K\!>\!8{,}192$ the V2 kernel
dispatches a split-$K$ variant that partitions the activation row
across multiple CTAs at a $\sim$10--20\% single-stream latency cost.
This is the default path for the linear-attention projections of
Qwen3.5-397B-A17B and is what carries Front~B (\S\ref{sec:exp-397b}).

\paragraph{V2a: an alternative for $K\!>\!8{,}192$.}
The V2a shared codebook library (\S\ref{sec:format-v2a}) takes
$\sim$1~KB of SMEM regardless of output dimension, extending the
single-CTA activation cache to $K\!\leq\!32{,}768$ on SM120/121 without
split-$K$. It is an alternative to the V2 split-$K$ path: V2a avoids
the partition overhead at the cost of slightly lower per-group codebook
precision (one library entry per group instead of one codebook per
channel). On the 397B Front~B workload, V2 with split-$K$ delivers
modestly higher single-stream throughput than V2a; we use V2 for the
headline numbers and report V2a as an option. The V2a kernel ships in
three dispatched variants (split-$N$ for the $M\!=\!1$ decode path;
split-$M$ for batched prefill at $M\!\geq\!16$; split-$K$ for
$K\!>\!8{,}192$), all sharing the V2 depack, gather, outlier-scatter,
GEMM skeleton.

\paragraph{Performance.}
On Qwen3.5-122B-A10B (RTX PRO 6000 Blackwell, TP=2), the V2 fused
kernel delivers 138~tok/s single-stream decode at $\sim$3.97 effective
bits (\S\ref{sec:experiments}). On Qwen3.5-397B-A17B at TP=2 with V2
plus split-$K$ dispatch under the H1.5 quality profile
(\S\ref{sec:exp-397b}) the kernel delivers 98.2~tok/s at medium
prompts and 100.9~tok/s at long prompts; the absolute throughput
reflects the model being $3.3\times$ larger by parameter count combined
with the split-$K$ partition overhead on linear-attention rows. At
$M\!=\!1$ the kernel remains memory-bandwidth bound; XFP's lower
effective bit width reads fewer weight bytes per token than INT4,
which is the structural source of its throughput advantage on the
122B and 397B headline workloads (\S\ref{sec:experiments}).

\section{Experiments}
\label{sec:experiments}

\subsection{Setup}

\paragraph{Platforms.} Two workstation-class Blackwell platforms:
(i)~\textbf{NVIDIA DGX Spark} (GB10, SM121, 120~GB unified memory),
single-GPU;
(ii)~\textbf{RTX PRO 6000 Blackwell} (SM120, $2\times$96~GB), with TP=2.
Native NVFP4 Tensor Core paths on SM120/121 required community fixes to
vLLM, FlashInfer, and CUTLASS dispatch and remain incomplete on
grouped-GEMM MoE workloads (\S\ref{sec:related}); XFP targets the bf16
SM path on this tier, which is what is reliably available.

\paragraph{Models.} GLM-4.7-Flash (30B, MoE-64, pure self-attention),
Qwen3.5-35B-A3B (35B, MoE-256), Qwen3.5-122B-A10B (122B, MoE-256, hybrid
GatedDeltaNet + self-attention) under V2; Qwen3.5-397B-A17B (397B,
MoE-512, hybrid GatedDeltaNet + self-attention) under V2a. All served
via quantize-on-load.

\paragraph{Evaluation.}
For sample-size-sensitive accuracy claims, we use
\textbf{lm-eval-harness 0.4.11}~\citep{eval-harness}: GSM8K, 5-shot,
\texttt{flexible-extract} and \texttt{strict-match}, 1{,}319 problems per
seed, 3 seeds per configuration.
For internal regression tracking and quick model-development feedback we
also use \texttt{bench.py}: deterministic (seed=42), 5 performance rounds,
50 GSM8K-style problems.

\paragraph{FP8 KV-cache and LM-head qualification.}
All XFP configurations use fp8 KV-cache and fp8 LM-head. To verify this
introduces no confounding accuracy loss, we measured the BF16 baseline of
Qwen3.5-35B-A3B under both bf16 and fp8 KV/LM-head pipelines: 76.12\% vs
76.35\% GSM8K strict-match ($\Delta = 0.2$~pp, within single-seed
variance). The fp8 infrastructure path is accuracy-neutral.

\subsection{Front A: Quality-Validated Deployment (V2)}
\label{sec:exp-frontA}

Front A reports paper-grade GSM8K under the V2 per-channel codebook on
models that fit comfortably within the workstation memory envelope.
The operative question is quality: how well does two-threshold
auto-select preserve accuracy against the best available 4-bit
calibration-aware baseline (AutoRound INT4 via Marlin)?

\begin{table}[t]
\centering
\caption{XFP auto-mode (V2, $\tau_\text{strict}\!=\!0.96$,
$\tau_\text{lazy}\!=\!0.93$). GSM8K via lm-eval-harness 0.4.11
(5-shot, 1{,}319 problems, 3 seeds, mean $\pm$ std). Throughput is
single-stream wall-clock decode at 1{,}500 output tokens.}
\label{tab:automode}
\scriptsize
\setlength{\tabcolsep}{4pt}
\begin{tabular}{@{}lccc@{}}
\toprule
Model & Eff.\ bits & tok/s & GSM8K strict \\
\midrule
Qwen 122B$^a$ & $\sim$3.97 & \textbf{138} & \textbf{94.49}$\pm$0.57 \\
Qwen 122B$^b$ & $\sim$3.97 & 110           & 94.62$\pm$0.67 \\
Qwen 35B$^b$  & $\sim$4.0  & 37            & 77.18$\pm$0.72 \\
\bottomrule
\end{tabular}
\end{table}

\noindent\footnotesize $^a$RTX PRO 6000 Blackwell, TP=2.
$^b$RTX PRO 6000 Blackwell, TP=1.\normalsize

Auto-mode adapts to model structure: on Qwen-122B-A10B routed experts
require XFP4 due to broader distributions ($\sim$3.97 eff.\ bits); on
Qwen-35B and GLM-4.7-Flash the same algorithm converges differently
according to per-layer cosine similarity (Table~\ref{tab:glm-auto}).

\subsubsection{Why Auto-Select Converges Differently per Class}
\label{sec:mixed}

Auto-mode's per-class bit-width choices reflect the underlying weight
distributions, not a hand-coded policy. Routed experts (95\% of MoE
parameters) tolerate XFP2 quantization with no measurable accuracy
degradation against an XFP4 baseline; the sensitivity bottleneck is in
attention and shared-expert projections. This is consistent with the
per-layer reconstruction data (Table~\ref{tab:outlier}): routed experts
show $\Delta\text{cos} < 0.001$ from outlier extraction---near-Gaussian
bulk distributions that a small codebook covers adequately. The
two-threshold auto-select (\S\ref{sec:encoding}) is the mechanism that
exploits this asymmetry without operator intervention.

\subsubsection{Outlier Extraction Ablation}

\begin{table}[t]
\centering
\caption{Outlier extraction effect (XFP3, $k$=4, GLM-4.7-Flash). MSE ratio
= MSE$_\text{bulk}$ / MSE$_\text{+outlier}$ at population median.}
\label{tab:outlier}
\scriptsize
\setlength{\tabcolsep}{4pt}
\begin{tabular}{@{}lcccc@{}}
\toprule
Layer type & $n$ & cos (bulk) & $\Delta$cos & MSE ratio \\
\midrule
attn\_kva       & 48 & 0.981 & +0.009 & 1.90$\times$ \\
attn\_kvb       & 48 & 0.983 & +0.004 & 1.24$\times$ \\
shared\_gate\_up & 94 & 0.981 & +0.002 & 1.13$\times$ \\
routed\_gate\_up & 96 & 0.982 & +0.0003 & 1.02$\times$ \\
routed\_down     & 96 & 0.982 & +0.0003 & 1.02$\times$ \\
\bottomrule
\end{tabular}
\end{table}

Outlier extraction is decisive for attention ($1.9\times$ MSE
improvement) but negligible for routed experts ($+0.0003$ cos).
Cross-model: comparing GLM-4.7-Flash with Qwen3.5-35B-A3B, GLM
concentrates catastrophic outliers (49$\sigma$) in
\texttt{kv\_a\_proj} while Qwen distributes moderate tails uniformly;
Qwen's routed experts respond $6.7\times$ more strongly to outlier
extraction. This is the data behind ``the algorithm is identical;
the profiles differ because distributions differ.''

\subsubsection{Decoder Throughput vs.\ Marlin INT4}

We use Marlin~\citep{frantar2024marlin} as the primary throughput
reference: it is the most optimized INT4 dequantization kernel
available on NVIDIA GPUs and serves AutoRound INT4
checkpoints~\citep{cheng2024autoround} that consistently match or
exceed NVFP4 in calibration-aware accuracy. AutoRound and XFP take
different methodological paths (uniform quantization with Hessian-aware
rounding vs.\ per-group learned codebooks with cosine-targeted
auto-select); the comparisons below are decoder-speed comparisons on
identical hardware, with accuracy reported alongside as a documentary
data point, not a verdict on either method.

\paragraph{RTX PRO 6000 Blackwell (SM120), TP=2.}
On a workstation-class dual-GPU node ($2\times$96~GB), the XFP V2
single-user reference kernel delivers 138~tok/s wall-clock single-stream
decode at full GSM8K quality (Table~\ref{tab:rtx-pro}). GPU utilization
during single-stream decode is 98\% on each device at 240~W, well below
the 600~W TDP---kernel-launch overhead in the single-stream path
dominates, leaving compute headroom that the present implementation
does not yet exploit. Massively concurrent batched serving is not a
goal of this work and would require a different kernel scheduling
strategy (see \S\ref{sec:limitations}).

\paragraph{DGX Spark (SM121), single-GPU.}
An earlier measurement on DGX Spark (SM121) shows the throughput
characteristic on a second platform: XFP outperforms Marlin INT4 by
16--18\% on long and medium decode (Table~\ref{tab:marlin-dgx}). The
architectural argument---fewer effective bits per weight reads fewer
bytes per token at $M\!=\!1$---holds across platforms.

\begin{table}[t]
\centering
\caption{XFP vs.\ Marlin INT4 on Qwen3.5-122B-A10B, DGX Spark, SM121
(\texttt{bench.py}, 50 problems).}
\label{tab:marlin-dgx}
\scriptsize
\setlength{\tabcolsep}{4pt}
\begin{tabular}{@{}lcccc@{}}
\toprule
 & long & medium & short & Math \\
\midrule
XFP auto    & \textbf{29.9} & \textbf{34.8} & 2.6 & 98\% \\
Marlin INT4 & 25.8          & 29.6          & 2.5 & 94\% \\
\midrule
$\Delta$    & +16\% & +18\% & $\sim$0 & --- \\
\bottomrule
\end{tabular}
\end{table}

The math difference (98\% vs 94\%) is observed on the \texttt{bench.py}
50-problem sample but not isolated from the AutoRound calibration
configuration of the Marlin checkpoint; the controlled comparison is
the throughput.

\paragraph{Paper-grade head-to-head on RTX PRO 6000.}
A controlled measurement at TP=1 on identical hardware
(Table~\ref{tab:marlin-rtx}) shows XFP at 110~tok/s vs Marlin INT4 at
73.6~tok/s ($+$49\% single-stream decode), with GSM8K strict-match of
94.62~$\pm$~0.67 vs 95.27~$\pm$~0.12. The $\sim$0.65~pp accuracy
difference is within the Lloyd-init jitter envelope ($\sim$0.67~pp
std on this scale); both formats land in the same 94--95\% band, with
the structural difference being throughput: XFP reads $\sim$3.97
effective bits per weight versus Marlin's 4.0, and the fused
depack--gather--scatter--GEMM kernel exploits this directly at the
memory-bandwidth limit.

\begin{table}[t]
\centering
\caption{XFP vs.\ Marlin INT4 (AutoRound) on Qwen3.5-122B-A10B,
RTX PRO 6000 Blackwell, SM120. lm-eval-harness 0.4.11, GSM8K
(5-shot, 1{,}319 problems, 3 seeds, mean $\pm$ std). Throughput is
single-stream wall-clock at 1{,}500 output tokens.}
\label{tab:marlin-rtx}
\scriptsize
\setlength{\tabcolsep}{4pt}
\begin{tabular}{@{}lccccc@{}}
\toprule
Format & TP & tok/s & flex & strict & Bits \\
\midrule
XFP V2 (auto) & 1 & \textbf{110} & 94.54$\pm$0.67 & 94.62$\pm$0.67 & $\sim$3.97 \\
XFP V2 (auto) & 2 & \textbf{138} & 94.41$\pm$0.57 & 94.49$\pm$0.57 & $\sim$3.97 \\
Marlin INT4   & 1 & 73.6         & 95.17$\pm$0.09 & \textbf{95.27}$\pm$0.12 & 4.0 \\
\bottomrule
\end{tabular}
\end{table}

\subsubsection{V2a Storage Mode (probes)}

The V2a storage mode (\S\ref{sec:format-v2a}) shares the auto-select
frontend and decode-kernel skeleton with V2 and differs only in
codebook layout (per-group library entry vs.\ per-channel codebook).
Bench-probe measurements ($n\!=\!50$ GSM8K, deterministic, seed=42,
RTX PRO 6000 TP=1) on three models show the speed characteristic
(Table~\ref{tab:v2a-probes}):

\begin{table}[t]
\centering
\caption{V2a storage-mode decode throughput probes,
\texttt{bench.py} ($n\!=\!50$, RTX PRO 6000 TP=1). Marlin
INT4 columns where available, on the same hardware in the
same session.}
\label{tab:v2a-probes}
\scriptsize
\setlength{\tabcolsep}{3.5pt}
\begin{tabular}{@{}lcccc@{}}
\toprule
Model & V2a & math & Marlin & vs.\ \\
      & tok/s & ($n\!=\!50$) & tok/s & Marlin \\
\midrule
Qwen3.5-35B-A3B   & 204.0          & 90\% & (no Q3.5) & --- \\
Qwen3.5-122B-A10B & \textbf{108.1} & 96\% & 73.6      & \textbf{+47\%} \\
GLM-4.7-Flash     & 118.1          & 68\% & 121.1     & $-$2.5\% \\
\bottomrule
\end{tabular}
\end{table}

On Qwen3.5-122B-A10B, V2a delivers $+$47\% throughput vs Marlin INT4
at comparable footprint; on GLM-4.7-Flash, the two are within
$\sim$2.5\% (XFP slightly slower, V2a math probe $+$4~pp on the
$n\!=\!50$ sample). Full-1{,}319 GSM8K runs under V2a on these models
are in the queue (\S\ref{sec:exp-397b} status note). The 35B
comparison is not currently fair: no Q3.5 INT4 Marlin checkpoint
exists at the time of measurement; the Q3.6 Marlin path is on a
different model variant.

\begin{table}[t]
\centering
\caption{XFP on Qwen3.5-122B-A10B, RTX PRO 6000 Blackwell, TP=2,
quantize-on-load. Decode throughput at 1{,}500-token output;
GSM8K via lm-eval-harness 0.4.11.}
\label{tab:rtx-pro}
\scriptsize
\setlength{\tabcolsep}{3pt}
\begin{tabular}{@{}lr@{}}
\toprule
Metric (single-stream reference) & Value \\
\midrule
Single-stream decode (wall)      & 138~tok/s \\
Engine sustained                 & 137~tok/s \\
\midrule
GSM8K flex.-extract$^*$          & \textbf{94.41}~$\pm$~0.57 \\
GSM8K strict-match$^*$           & \textbf{94.49}~$\pm$~0.57 \\
\midrule
Effective bits                   & $\sim$3.97 \\
VRAM / GPU                       & 57 / 96~GiB \\
GPU util.\ (single-stream)       & 98\% / 98\% \\
GPU power (single-stream)        & 240 / 240~W \\
\bottomrule
\end{tabular}\\[2pt]
{\scriptsize $^*$3 seeds, 1{,}319 problems/seed, mean $\pm$ std (\%). All
numbers from the single-user reference kernel; concurrent / batched
serving is out of scope (\S\ref{sec:limitations}).}
\end{table}

\paragraph{Cross-platform consistency.}
The same V2 auto-mode quantization profile ($\sim$3.97 eff.\ bits)
generalizes from a single-GPU 120~GB unified-memory device (DGX Spark) to
a dual-GPU discrete TP=2 setup (RTX PRO 6000) without code changes.
Together these two platforms span the workstation-class hardware tier
where native NVFP4 Tensor Core support remains community-patched and
incomplete on grouped-GEMM MoE workloads---and where, unlike server-class
A100/H100 deployments, memory-bound 4-bit kernels matter most. XFP
places this tier at top-of-class for 100B+ MoE inference.

\begin{figure}[t]
\centering
\begin{tikzpicture}
\begin{axis}[
    ybar,
    width=0.95\columnwidth, height=4.8cm,
    bar width=18pt,
    ylabel={tok/s},
    symbolic x coords={XFP TP=1, Marlin TP=1, XFP TP=2},
    xtick=data,
    x tick label style={font=\scriptsize},
    y tick label style={font=\scriptsize},
    ylabel style={font=\small},
    ymin=0, ymax=170,
    nodes near coords,
    every node near coord/.append style={font=\tiny},
    grid=major, grid style={gray!15},
    enlarge x limits=0.18,
]
\addplot[fill=xfpblue!70, draw=xfpblue] coordinates
    {(XFP TP=1, 110) (Marlin TP=1, 73.6) (XFP TP=2, 138)};
\end{axis}
\end{tikzpicture}
\caption{\textbf{Single-stream decode throughput on Qwen3.5-122B-A10B},
RTX PRO 6000 Blackwell (SM120), 1{,}500-token output. At identical TP=1
single-stream (the regime this work targets), XFP is $+$49\% faster
than Marlin INT4 (AutoRound); TP=2 extends this to $+$87\%. Both XFP
and Marlin are memory-bandwidth-bound at $M\!=\!1$; XFP reads $\sim$3.97
effective bits per weight versus Marlin's 4.0. Concurrent / batched
serving is out of scope (\S\ref{sec:limitations}).}
\label{fig:throughput}
\end{figure}
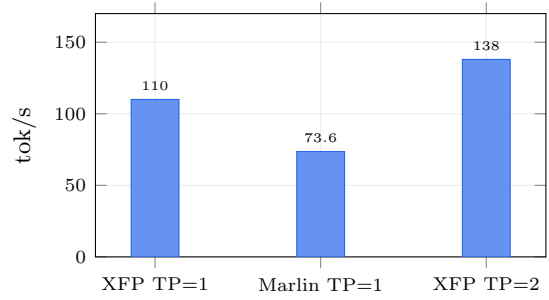

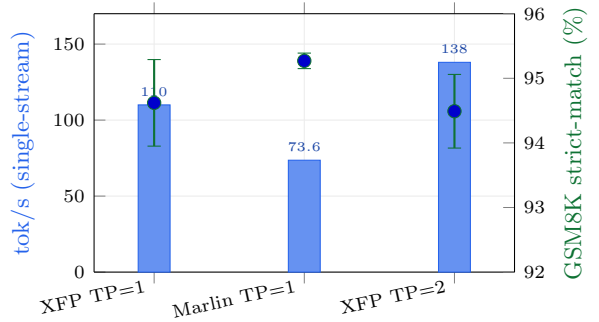
\begin{figure}[t]
\centering
\begin{tikzpicture}
\begin{axis}[
    ybar,
    width=0.90\columnwidth, height=5cm,
    bar width=12pt,
    ylabel={tok/s (single-stream)},
    axis y line*=left,
    symbolic x coords={XFP TP=1, Marlin TP=1, XFP TP=2},
    xtick=data,
    x tick label style={font=\scriptsize, rotate=12, anchor=east},
    y tick label style={font=\scriptsize},
    ylabel style={font=\small, color=xfpblue},
    ymin=0, ymax=170,
    nodes near coords,
    every node near coord/.append style={font=\tiny, color=xfpblue!70!black},
    grid=major, grid style={gray!15},
    enlarge x limits=0.20,
    name=throughputaxis,
]
\addplot[fill=xfpblue!70, draw=xfpblue] coordinates
    {(XFP TP=1, 110) (Marlin TP=1, 73.6) (XFP TP=2, 138)};
\end{axis}
\begin{axis}[
    width=0.90\columnwidth, height=5cm,
    only marks,
    mark=*, mark size=2.5pt, mark options={fill=xfpgreen, draw=xfpgreen!70!black},
    ylabel={GSM8K strict-match (\%)},
    axis y line*=right,
    axis x line=none,
    symbolic x coords={XFP TP=1, Marlin TP=1, XFP TP=2},
    xtick=data,
    ymin=92, ymax=96,
    y tick label style={font=\scriptsize},
    ylabel style={font=\small, color=xfpgreen!70!black},
    enlarge x limits=0.20,
]
\addplot+[error bars/.cd, y dir=both, y explicit,
    error bar style={thick, xfpgreen!70!black}]
    coordinates {
        (XFP TP=1, 94.62) +- (0, 0.67)
        (Marlin TP=1, 95.27) +- (0, 0.12)
        (XFP TP=2, 94.49) +- (0, 0.57)
    };
\end{axis}
\end{tikzpicture}
\caption{\textbf{XFP vs.\ Marlin INT4 on Qwen3.5-122B-A10B}, RTX PRO
6000 Blackwell. Bars (left): single-stream tok/s. Markers (right): GSM8K
strict-match (3 seeds, mean $\pm$ std). At TP=1, XFP is 49\% faster at
$-$0.65~pp accuracy (within seed-variance).}
\label{fig:gsm8k}
\end{figure}

\subsection{Front B: The H-Process --- Constrained Compression on a 397B Model}
\label{sec:exp-397b}

Qwen3.5-397B-A17B is a hybrid linear-/self-attention MoE with 512
routed experts per layer, 60 layers, and an unquantized BF16 footprint
of $\sim$795~GB---one order of magnitude above the 35B/122B platforms
of Front~A. At any conventional 4-bit format ($\sim$200~GB after
quantization plus KV-cache, activations, and library buffers), the
model does not fit on $2\times$96~GB. Front~B reports the
\emph{H-Process}: a quality-driven iteration over the two cosine
thresholds (\S\ref{sec:encoding}) that finds the operating point at
which the model just fits while still producing sensible output.

\paragraph{Three constraints define the H-Process search space.}
The operator-facing constraint is the pair
$(\tau_\text{strict}, \tau_\text{lazy})$: bit widths are derived from
these, not chosen directly. Two hard boundaries flank the operating
window:

\begin{itemize}[leftmargin=1.2em,itemsep=1pt,topsep=2pt]
\item \textbf{OOM at quantize-on-load.} High thresholds force more
  routed-expert groups into xfp4, raising the Lloyd-iteration spike
  past the 96~GB-per-GPU envelope. The boundary is set by the
  encoding spike, not the steady-state serving footprint:
  serving-side reductions (KV-cache, \texttt{max\_model\_len},
  \texttt{gpu\_memory\_utilization}) do not help, because the spike
  occurs before serving allocation (Table~\ref{tab:cos-sweep},
  H1.7 entry).
\item \textbf{Garbage output.} Low thresholds force most routed
  experts into xfp2; bit-width reconstruction passes the cosine gate
  numerically, but generation degenerates to repeated tokens despite
  apparently valid arithmetic. Cosine similarity is a chosen
  \emph{indicator} for steering bit-width selection, not a quality
  guarantee. The garbage boundary exists strictly inside the
  cos-passes region and is the empirical reminder that the operating
  envelope is verified by benches, not by the indicator alone.
\end{itemize}

The H-Process iterates $(\tau_\text{strict}, \tau_\text{lazy})$ from
aggressive toward conservative settings until both boundaries are clear
and the resulting model serves with acceptable bench output.

\paragraph{Hardware and serving setup.}
$2\times$RTX PRO 6000 Blackwell (SM120, 96~GB each), TP=2, BF16
KV-cache (FlashInfer FP8-KV is incompatible with the model's
$\text{head\_dim}\!=\!256$ projections; the Triton attention backend
with BF16 KV is the path that runs).
\texttt{XFP\_GROUP\_SIZE}=128, \texttt{XFP\_MOE\_LLOYD\_ITERS}=20,
LM-head FP8, \texttt{max\_model\_len}=65{,}536, custom all-reduce
disabled (a known SM120 issue at TP=2). The engine is V2 (per-group
Lloyd codebook) with split-$K$ dispatch on the linear-attention
projections that exceed $K\!=\!8192$; under the V2a storage mode
(\S\ref{sec:format-v2a}) we observe equivalent functionality with
slightly lower throughput, and use V2 for the headline numbers below.

\paragraph{The quality gradient.}
With the candidate set effectively $\{2,4\}$ (xfp3 is rare on this
model under either storage mode, see \S\ref{sec:format-v2a}), the
two-threshold split (\S\ref{sec:encoding}) controls memory through
the lazy-vs-strict gap. Table~\ref{tab:cos-sweep} reports five
operating points; the strict floor is held at
$\tau_\text{strict}\!=\!0.96$ throughout. The dependence on
$\tau_\text{lazy}$ is steep and sharply non-linear: at $0.80$ the
routed-expert path collapses to near-100\% xfp2 and math output
degenerates to repeated tokens (cosine passes, benches do not); at
$0.935$ already enough routed-expert groups escalate to xfp4 that the
Lloyd-iteration spike exceeds the 96~GB envelope. The OOM boundary
is set by the encoding spike, not the steady-state serving footprint:
H1.7 ($\tau_\text{lazy}\!=\!0.935$) failed across three load attempts
even with aggressive serving-side reduction
(\texttt{gpu\_mem\_util}~0.97, KV-cache 2~GB, \texttt{max\_model\_len}~8192),
because the spike occurs before any serving allocation. Between the
two boundaries the window is narrow: $\tau_\text{lazy}\!=\!0.92$
yields 52\% GSM8K strict-match at $n\!=\!50$ and 72~GiB/GPU steady
state; raising to $0.93$ adds 11~GiB/GPU and shifts the
$n\!=\!50$ probe by $+8$~pp. On the H1.5 configuration
($\tau_\text{lazy}\!=\!0.93$, $\sim$3.4 effective bits) the
full-1{,}319 single-seed GSM8K strict-match reaches \textbf{66.72\%}
at 100.9~tok/s long-output decode.

\begin{table}[t]
\centering
\caption{H-Process operating points on Qwen3.5-397B-A17B,
$2\times$RTX PRO 6000 TP=2, engine V2. Steady = post-load
steady-state VRAM per GPU; Spike = max during quantize-on-load.
Probe accuracy on \texttt{bench.py} ($n\!=\!50$ GSM8K, deterministic,
seed=42); H1.5 also reports full-1{,}319 single-seed paper-grade
GSM8K. H1.7 was attempted three times with successively
more aggressive serving-side reductions (see text); all three
attempts failed during the encoding spike. Multi-seed H1.5
evaluation is in progress.}
\label{tab:cos-sweep}
\scriptsize
\setlength{\tabcolsep}{2.5pt}
\begin{tabular}{@{}lccccccc@{}}
\toprule
Var. & $\tau_\text{strict}$ & $\tau_\text{lazy}$ & Steady & Spike & Eff & probe & full-1319 \\
     &                       &                     & GiB    & GiB   & bits & ($n\!=\!50$) & (1 seed) \\
\midrule
G    & 0.92  & 0.80  & 70 & 96 & $\sim$2.3 & garbage & --- \\
H1   & 0.96  & 0.92  & 72 & 96 & $\sim$2.6 & 0.52$\pm$0.07 & --- \\
\textbf{H1.5} & 0.96 & 0.93 & 83 & 97 & $\sim$3.4 & 0.60$\pm$0.07 & \textbf{0.6672} \\
H1.7 & 0.96  & 0.935 & \multicolumn{2}{c}{OOM (3$\times$)} & --- & --- & --- \\
H    & 0.96  & 0.94  & \multicolumn{2}{c}{OOM} & --- & --- & --- \\
\bottomrule
\end{tabular}
\end{table}

\paragraph{Per-class bit-width assignment under H1.5.}
Auto-select picks xfp4 for 100\% of self-attention, linear-attention
and shared-expert projections, and xfp2 for the majority of
routed-expert groups; xfp4 escalations on the routed path
concentrate in layers~1--3 (the initial GatedDeltaNet transition) and
layer~13. The pattern is stable across runs and is consistent with the
per-component outlier-extraction sensitivity reported in
Table~\ref{tab:outlier}: routed experts have near-Gaussian bulk
distributions that admit aggressive quantization, attention does not.

\paragraph{Throughput.}
Single-stream wall-clock decode on H1.5: 73.4~tok/s at short outputs,
98.2~tok/s at medium, \textbf{100.9~tok/s at long outputs}
($n\!=\!5$ runs, \texttt{temp}=0, $1{,}500$-token outputs).
H1 throughput on the same setup is approximately
65--97~tok/s depending on output length.

\paragraph{An alternative path: routed-expert reduction.}
A different family of methods addresses the same memory constraint
by reducing the routed-expert population at fixed 4-bit precision.
We compare against the REAP-it-Yourself (RIY) workflow on the same
model and hardware, applying AutoRound INT4 with REAP-style
calibration-aware expert removal (\S\ref{sec:related}).
Table~\ref{tab:int4-riy} reports two RIY operating points alongside
the H-Process. The two approaches reach the same envelope along
different axes: XFP+H reduces per-weight bits while retaining all
experts; RIY retains 4-bit precision while removing routed experts.
Both are legitimate responses to the constraint; the comparison
documents how the axes trade off on this particular model.

\begin{table}[t]
\centering
\caption{Qwen3.5-397B-A17B on $2\times$RTX PRO 6000 TP=2:
H-Process versus AutoRound INT4 with REAP-style routed-expert
pruning. INT4 numbers from earlier deployment runs of the same
model on the same hardware; GSM8K via \texttt{bench.py} $n\!=\!50$
unless otherwise noted.}
\label{tab:int4-riy}
\scriptsize
\setlength{\tabcolsep}{3pt}
\begin{tabular}{@{}lccccc@{}}
\toprule
Format & Prune & Steady & GSM8K & $n$ & tok/s \\
       &       & GiB/GPU & strict &     & long \\
\midrule
INT4 + RIY        & $-$36\% & $\sim$75 & 33\%             & 50       & $\sim$50 \\
INT4 + RIY        & $-$24\% & $\sim$80 & 44\%             & 50       & $\sim$50 \\
XFP H1            & 0\%     & 72       & 52\%             & 50       & 65 \\
\textbf{XFP H1.5} & 0\%     & 83       & \textbf{66.72\%} & 1319$^*$ & \textbf{100.9} \\
\bottomrule
\end{tabular}\\[2pt]
{\scriptsize INT4 = AutoRound INT4 served via Marlin.
$^*$Single seed; multi-seed evaluation in progress.}
\end{table}

On this model, the H-Process operates at a lower memory footprint
than RIY-24\% (72 vs 80~GiB/GPU at H1; 83 at H1.5), with the
$n\!=\!50$ probe favoring XFP at both operating points and the
full-1{,}319 H1.5 measurement extending the gap further. On
throughput, the XFP fused decode kernel reads fewer bytes per token
than INT4 at $M\!=\!1$ and exploits this directly at the
memory-bandwidth limit; this is the structural reason for the speed
difference, not a Marlin-specific shortfall (\S\ref{sec:related},
Marlin is the most optimized INT4 dequantization kernel available and
serves as our throughput reference).

\paragraph{Hardware projection.}
Reserving 8~GiB for KV-cache and 10~GiB for activations and compile
buffers, the H1 steady state fits comfortably on $2\times$96~GB
(6~GiB/GPU headroom); H1.5 is marginal (effectively zero headroom).
H100~80~GB SXM is not viable at TP=2 for either profile; H200~141~GB
TP=2 has $\sim$51~GiB headroom and should serve H1.5 comfortably.
Single-device deployment on the DGX Spark (120~GB unified) is not
viable at the present effective bit widths without further reduction.

\paragraph{Status.}
The H1.5 GSM8K strict-match (66.72\%, full-1{,}319) is a single-seed
measurement at submission time; multi-seed evaluation is in progress
and will replace the single-seed value in the final version. H1
paper-grade evaluation and a full-1{,}319 measurement on the V2a
storage mode are also in the queue.

\section{Discussion}
\label{sec:discussion}

\subsection{Two Quality Floors as Operator Knob}

The two-threshold split ($\tau_\text{strict}\!=\!0.96$ for
attention/shared, $\tau_\text{lazy}\!=\!0.93$ for routed-expert MoE,
\S\ref{sec:encoding}) collapses the per-layer auto-select into a single
operator-facing parameter: the threshold gap. On the 397B sweep
(Table~\ref{tab:cos-sweep}), each +0.01 to $\tau_\text{lazy}$ shifts
roughly one xfp2$\to$xfp4 transition per routed-expert layer, costing
$\sim$11~GiB/GPU steady-state and yielding $+8$~pp GSM8K. The same
two-threshold mechanism applied to 122B under V2 reproduces the
single-threshold $\tau\!=\!0.98$ behaviour: with $\mathcal{N}\!=\!\{2,3,4\}$
the intermediate option $N\!=\!3$ absorbs the elasticity, and the lazy
floor is rarely binding.

\subsection{Storage Tradeoff: Bits vs.\ Outliers}

On the linear path, each outlier costs 18 bytes. The break-even between
XFP3 with elevated outlier fraction and XFP4 at 2\% cap is:
$0.375 + 18y = 0.5 + 0.02 \times 18 \implies y = 2.7\%$. The default 2\%
cap sits near this break-even---no hidden win from aggressive outlier
extraction.

\subsection{The Minimalist Device Dilemma}
\label{sec:minimalist}

The $K\!\leq\!8192$ bound under V2 is not driven by algorithmic
constraints---the same kernel design works at $K\!\leq\!32{,}768$ on
data-center silicon. It is a direct consequence of how shared-memory
capacity is segmented across the Blackwell generation:

\begin{table}[ht]
\centering
\caption{Per-CTA opt-in shared-memory limit across NVIDIA SMs.}
\label{tab:smem-segmentation}
\scriptsize
\setlength{\tabcolsep}{4pt}
\begin{tabular}{@{}llr@{}}
\toprule
SM tier & Hardware & SMEM/CTA \\
\midrule
SM90  (Hopper d/c)         & H100, H200       & 228~KB \\
SM100 (Blackwell d/c)      & B100, B200       & 228~KB \\
SM120 (Blackwell wkst.)    & RTX PRO 6000     & \textbf{99~KB} \\
SM121 (Blackwell SoC)      & DGX Spark (GB10) & \textbf{99~KB} \\
\bottomrule
\end{tabular}
\end{table}

The same building blocks are present on each tier; the per-CTA SMEM
ceiling on the workstation tier is less than half that of data-center
silicon. For memory-bound 4-bit GEMM kernels with activation-row caching,
the largest contiguous $K$ drops from $\sim$32K to $\sim$8K under V2's
per-channel codebook---a $4\times$ regression on the dimension governing
single-user decode latency. V2a's shared codebook library lifts the
bound back to $K\!\leq\!32{,}768$ on the workstation tier
(\S\ref{sec:format-v2a},~\S\ref{sec:kernel}), restoring the data-center
operating envelope at the cost of the per-channel codebook precision.
For V2 users on the workstation tier, a split-$K$ extension lifts the
$K\!\leq\!8192$ bound at $\sim$10--20\% latency cost
(\S\ref{sec:limitations}).

\subsection{Lloyd Non-Determinism}

Quantize-on-load does not produce bit-identical weights across runs.
On Qwen-35B, identical bit-width decisions yield $\pm$8pp math variance
due to Lloyd initialization jitter. Weight caching eliminates run-to-run
variance after the first load.

\subsection{Limitations and Scope}
\label{sec:limitations}

\begin{enumerate}[leftmargin=*,itemsep=2pt]
\item \textbf{Single-user reference implementation.} The XFP fused
  decode kernel presented here is optimized for low-latency single-stream
  inference. At concurrency $\geq$8, per-request throughput collapses
  because the per-call kernel-launch overhead is not amortized across
  the batch. A high-throughput batched-serving variant is
  planned future work and is explicitly out of scope for this paper.
  All accuracy and throughput numbers reported here are from the
  single-user reference path.
\item \textbf{$K\!\leq\!8192$ bound under V2 on the workstation tier.}
  The V2 fused decode kernel caches the full activation row in shared
  memory ($\sim$24~KB at $K\!=\!8192$ in bf16; see \S\ref{sec:kernel}).
  On the target workstation hardware (DGX Spark, RTX PRO 6000 Blackwell;
  both 99~KB opt-in SMEM per CTA) this bound covers the Qwen3.5 family
  but excludes models with wider intermediate dimensions. V2a lifts
  this bound to $K\!\leq\!32{,}768$ via the shared-library codebook
  (\S\ref{sec:format-v2a}); a split-$K$ V2 variant is also available
  at a 10--20\% single-stream latency cost.
\item \textbf{No Hessian feedback.} Hessian-weighted Lloyd is a planned
  extension.
\item \textbf{Lloyd non-determinism.} $\pm$8pp math variance on 50-problem
  benchmarks under \texttt{bench.py} probing (each problem = 2pp); under
  paper-grade lm-eval-harness with 1{,}319 problems and 3 seeds the
  cross-seed std drops to 0.57\% on GSM8K (Table~\ref{tab:rtx-pro}).
\item \textbf{Evaluation scope.} GSM8K under lm-eval-harness is the
  primary accuracy metric. MMLU and additional reasoning suites are
  natural extensions for future work.
\item \textbf{Hardware.} Validated on two workstation-class platforms
  (DGX Spark SM121, RTX PRO 6000 Blackwell SM120). Native NVFP4 Tensor
  Core paths on these SMs required community fixes to vLLM, FlashInfer,
  and CUTLASS dispatch (\S\ref{sec:related}); server-class A100/H100
  evaluation is not part of this work as it would not meaningfully test
  the memory-bound regime XFP targets.
\item \textbf{Auto-mode range.} V2 tests $N \in \{2,3,4\}$; V2a tests
  $N \in \{2,4\}$ (the $g\!=\!128$ packing constraint excludes
  $N\!=\!3$). A $g\!=\!80$ V2a path supporting $N\!=\!3$ is deferred.
\end{enumerate}

\section{Conclusion}

XFP is a quality-targeted, adaptive weight quantizer that inverts the
conventional workflow: the operator specifies a cosine-similarity
floor; XFP picks bit widths, outlier budgets, and packing per layer
to meet it. The format and its supporting auto-select frontend, fused
decode kernel, and two storage modes (per-channel V2 and shared-library
V2a) are not the contribution by themselves---they are means to an
end. The end is appliance-class deployment: large language models
serving on workstation hardware, without datacenter backends and
without NVFP4-accelerated Tensor Core paths that arrived late and
remain incomplete on this tier.

Four results support this trajectory:
(1)~the same auto-select frontend drives both codebook storage modes,
covering the 30B--397B parameter range on the same workstation
platforms;
(2)~auto-mode converges to $\sim$3.97 effective bits on
Qwen3.5-122B-A10B and to $\sim$2.6--3.4 on Qwen3.5-397B-A17B
depending on the H-Process operating point;
(3)~XFP on Qwen3.5-122B-A10B (RTX PRO 6000, TP=2) under V2 reaches
138~tok/s single-stream decode at \textbf{94.49\% GSM8K strict-match}
(3 seeds, $n\!=\!3{,}957$ problems)---paper-grade accuracy preservation
on workstation-class dual-GPU hardware;
(4)~the H-Process on Qwen3.5-397B-A17B (512 routed experts/layer)
serves the full expert population in $2\times96$~GB at $\sim$3.4
effective bits and 100.9~tok/s long-output decode, with 66.72\%
GSM8K strict-match on the full 1{,}319-problem set (single seed,
multi-seed in progress).

\paragraph{The H-Process closes the loop to the motivation.}
The H-Process (\S\ref{sec:exp-397b}) is the application that makes
the format into a process. The motivation was appliance deployment
of models that do not fit; the H-Process is the quality-driven
iteration over the two cosine thresholds that finds the operating
point at which a given model just fits on a given hardware envelope
while still producing sensible output. Three constraints define its
search space: the operator-set $(\tau_\text{strict}, \tau_\text{lazy})$
pair, the OOM boundary at quantize-on-load, and the garbage boundary
in generation. The garbage boundary is the empirical reminder that
cosine similarity is a chosen \emph{indicator} for steering bit-width
selection---a much better indicator than bit width alone, but not a
guarantee. Quality is verified by benches; cosine steers. The
H-Process puts both in the operator's hands as explicit knobs, where
fixed-bit-width formats have neither.

\paragraph{A different family of methods.}
XFP belongs to a family of per-group codebook approaches; the
dominant 4-bit deployment formats are uniform quantizers with
calibration-aware corrections (AutoRound, AWQ, GPTQ) decoded by
Marlin or NVFP4-class kernels. These are different methodological
choices for the same goal, with different strengths. XFP requires no
Hessian, no calibration data, and no specialized hardware---and
provides the operator a tunable quality knob that uniform formats do
not. The trade-off is the per-group codebook overhead and the
encoding-time Lloyd iterations. On the deployments measured here,
the trade-off favors XFP at the workstation tier; on other workloads
it may not. The choice is a deployment decision, not a benchmark
verdict. Code is available at the project repository linked above.

\appendix
\onecolumn

\section{Per-Expert Reconstruction Tables}
\label{app:reconstruction}

All tables report per-layer metrics aggregated across $n$ layers per
component type, at XFP3 ($N\!=\!3$) with outlier threshold $k\!=\!4.0$.

\textbf{Column definitions:}
\textbf{cos bulk}---cosine similarity of codebook-only reconstruction (no
outlier path);
\textbf{cos outlier}---cosine similarity with sparse fp16 outlier
extraction enabled;
\textbf{$\Delta$cos mean / max}---improvement from outlier extraction
across the layer population;
\textbf{MSE p50 / p90}---ratio of MSE$_\text{bulk-only}$ to
MSE$_\text{with-outliers}$ at the 50th and 90th percentile of the layer
population (values $>1\times$ indicate outlier extraction reduces error;
e.g.\ $1.90\times$ means the bulk-only MSE is $1.9\times$ larger).

\subsection{GLM-4.7-Flash (XFP3, $k$=4)}

\begin{table}[ht]
\centering
\small
\begin{tabular}{@{}lccccccr@{}}
\toprule
Type & $n$ & cos bulk & cos outlier & $\Delta$cos mean & $\Delta$cos max & MSE p50 & MSE p90 \\
\midrule
attn\_kva       & 48 & 0.98070 & 0.98973 & +0.00903 & +0.01495 & 1.90$\times$ & 2.26$\times$ \\
attn\_kvb       & 48 & 0.98294 & 0.98654 & +0.00360 & +0.01051 & 1.24$\times$ & 1.36$\times$ \\
shared\_gate\_up & 94 & 0.98114 & 0.98356 & +0.00242 & +0.00600 & 1.13$\times$ & 1.23$\times$ \\
shared\_down    & 47 & 0.97964 & 0.98195 & +0.00231 & +0.00828 & 1.07$\times$ & 1.27$\times$ \\
attn\_o         & 48 & 0.98212 & 0.98391 & +0.00179 & +0.00627 & 1.06$\times$ & 1.14$\times$ \\
attn\_qb        & 48 & 0.98173 & 0.98336 & +0.00163 & +0.00964 & 1.06$\times$ & 1.17$\times$ \\
attn\_qa        & 48 & 0.98191 & 0.98264 & +0.00073 & +0.00303 & 1.03$\times$ & 1.08$\times$ \\
routed\_down    & 96 & 0.98238 & 0.98265 & +0.00027 & +0.00053 & 1.02$\times$ & 1.02$\times$ \\
routed\_gate\_up & 96 & 0.98237 & 0.98265 & +0.00028 & +0.00050 & 1.02$\times$ & 1.02$\times$ \\
\bottomrule
\end{tabular}
\caption{GLM-4.7-Flash per-expert reconstruction statistics.}
\label{tab:glm-recon}
\end{table}

\subsection{Qwen3.5-35B-A3B (XFP3, $k$=4)}

\begin{table}[ht]
\centering
\small
\begin{tabular}{@{}lccccccr@{}}
\toprule
Type & $n$ & cos bulk & cos outlier & $\Delta$cos mean & $\Delta$cos max & MSE p50 & MSE p90 \\
\midrule
shared\_gate\_up  & 96 & 0.96555 & 0.98279 & +0.01724 & +0.06805 & 1.39$\times$ & 4.70$\times$ \\
shared\_down     & 41 & 0.97757 & 0.98220 & +0.00463 & +0.01780 & 1.24$\times$ & 1.29$\times$ \\
attn\_other      & 33 & 0.97900 & 0.98431 & +0.00531 & +0.02224 & 1.21$\times$ & 1.46$\times$ \\
attn\_o          & 11 & 0.97909 & 0.98302 & +0.00393 & +0.01109 & 1.17$\times$ & 1.28$\times$ \\
routed\_down     & 96 & 0.98025 & 0.98207 & +0.00182 & +0.01174 & 1.07$\times$ & 1.20$\times$ \\
routed\_gate\_up  & 96 & 0.98059 & 0.98238 & +0.00180 & +0.00963 & 1.08$\times$ & 1.16$\times$ \\
\bottomrule
\end{tabular}
\caption{Qwen3.5-35B-A3B per-expert reconstruction statistics.}
\label{tab:qwen-recon}
\end{table}

\section{Codebook Analysis Detail}
\label{app:codebook}

\subsection{Cross-Model Distribution Profiles}

\begin{table}[ht]
\centering
\small
\begin{tabular}{@{}lcccc@{}}
\toprule
Component & GLM 3$\sigma$\% & Qwen 3$\sigma$\% & GLM max$|w|$ & Qwen max$|w|$ \\
\midrule
attn\_k         & 1.48 & 0.94 & 1.63 (49$\sigma$)  & 0.22 (14$\sigma$) \\
attn\_v         & ---  & 0.82 & ---           & 0.20 \\
dense\_mlp      & 0.40 & 1.19 & 0.94          & 0.32 \\
routed\_down    & 0.28 & 0.40 & 0.26          & 0.32 \\
shared\_gate\_up & 0.54 & 1.02 & 0.49          & 0.13 \\
\bottomrule
\end{tabular}
\caption{Weight distribution profiles: GLM-4.7-Flash vs.\ Qwen3.5-35B-A3B.}
\label{tab:distributions}
\end{table}

\section{MoE Sampling Validation}
\label{app:sampling}

Offline validation of the 4-expert sampling approximation for auto-select
bit-width decisions, comparing against full-population auto-select
(Table~\ref{tab:sampling-offline}), and live end-to-end corroboration of
the resulting bit-width decisions on running models
(Table~\ref{tab:sampling-live}):

\begin{table}[!ht]
\centering
\small
\begin{tabular}{@{}lccccl@{}}
\toprule
Model & Experts & MoE blocks & 4-sample $\neq$ full & \% & Failure mode \\
\midrule
GLM-4.7-Flash    & 64  & 94 & 0  & 0\%    & --- \\
Qwen3.5-35B-A3B  & 256 & 80 & 0  & 0\%    & --- \\
Qwen3.5-122B-A10B & 256 & 96 & 11 & 11.5\% & 10$\times$ under, 1$\times$ over \\
\bottomrule
\end{tabular}
\caption{Offline bit-width agreement: 4-expert sample vs.\ full population.}
\label{tab:sampling-offline}
\end{table}

\begin{table}[!ht]
\centering
\small
\begin{tabular}{@{}lccc@{}}
\toprule
Model & 4-sample tok/s / Math & All-experts tok/s / Math & $\Delta$ \\
\midrule
GLM-4.7-Flash    & 59.8 / 52\% & 59.5 / 52\% & Byte-identical \\
Qwen3.5-35B-A3B  & 55.4 / 82\% & 54.7 / 90\% & Bits identical; $\Delta$Math = Lloyd noise \\
Qwen3.5-122B     & 29.9 / 98\% & (UMA-blocked) & 2 layers flip as predicted \\
\bottomrule
\end{tabular}
\caption{Live E2E corroboration: 4-expert sample vs.\ all-experts.}
\label{tab:sampling-live}
\end{table}

Across 178 validated MoE blocks (all GLM + 35B + 32 live layers on 122B),
live bit-width decisions agree 100\% with offline predictions. The $\pm$8pp
math variance on Qwen-35B reflects Lloyd non-determinism (identical bit
decisions, different codebook centroids), not a sampling effect.

\twocolumn

\bibliographystyle{plainnat}

\end{document}